\documentclass[conference]{IEEEtran}
\IEEEoverridecommandlockouts
\usepackage{cite}
\usepackage{amsmath,amssymb,amsfonts}
\usepackage{graphicx}
\usepackage{textcomp}
\usepackage{xcolor}
\usepackage{booktabs}
\usepackage{algorithm}
\usepackage{algorithmicx}
\usepackage{algpseudocode}
\usepackage{multirow}
\usepackage{caption} 
\usepackage{subcaption}
\usepackage{hyperref}

\def\BibTeX{{\rm B\kern-.05em{\sc i\kern-.025em b}\kern-.08em
    T\kern-.1667em\lower.7ex\hbox{E}\kern-.125emX}}
\begin{document}

\title{High-Quality Pseudo-Label Generation Based on Visual Prompt Assisted Cloud Model Update}

\author{
\IEEEauthorblockN{
Xinrun Xu\textsuperscript{1,2}$^\text{*}$, 
Qiuhong Zhang\textsuperscript{1,2}$^\text{*}$, 
Jianwen Yang\textsuperscript{1,2},
Zhanbiao Lian\textsuperscript{1,2}, 
Jin Yan\textsuperscript{1,2},
Zhiming Ding\textsuperscript{2}$^\text{\dag}$,
Shan Jiang\textsuperscript{3}$^\text{\dag}$
}
\IEEEauthorblockA{
\textsuperscript{1}{University of Chinese Academy of Sciences}, Beijing, China\\
\textsuperscript{2}{Institute of Software, Chinese Academy of Sciences}, Beijing, China \\
\textsuperscript{3}{Advanced Institute of Big Data, Beijing, China}
}
Email:\{xuxinrun20, zhangqiuhong22, yangjianwen18, lianzhanbiao21, yvette.yan\}@mails.ucas.ac.cn,\\
zhiming@iscas.ac.cn,
jiangshan@alumni.nudt.edu.cn
}
\maketitle
\renewcommand{\thefootnote}{\fnsymbol{footnote}}
\footnotetext[1]{Equal Contribution.} 
\footnotetext[2]{Corresponding Author.}
\footnotetext[3]{This work was supported by the National Key R\&D Program of China (No. 2022YFF0503900).}
\renewcommand{\thefootnote}{\arabic{footnote}}


\begin{abstract}
Generating high-quality pseudo-labels on the cloud side is crucial for cloud-edge collaborative object detection, especially in dynamic traffic monitoring scenarios where the target data distribution continuously evolves. Existing methods often assume a perfectly reliable cloud model, neglecting the potential for errors in the cloud's predictions, or employ simple adaptation techniques that struggle to handle complex distribution shifts. This paper proposes a novel Cloud-Adaptive High-Quality Pseudo-label generation algorithm (CA-HQP) that addresses these limitations by incorporating a learnable Visual Prompt Generator (VPG) and a dual feature alignment strategy into the cloud model updating process. The VPG enables parameter-efficient adaptation of the large pre-trained cloud model by injecting task-specific visual prompts into the model's input, enhancing its flexibility without extensive fine-tuning. To mitigate domain discrepancies, CA-HQP introduces two complementary feature alignment techniques: a global Domain Query Feature Alignment (DQFA) that captures scene-level distribution shifts and a fine-grained Temporal Instance-Aware Feature Embedding Alignment (TIAFA) that addresses instance-level variations. Extensive experiments on the Bellevue traffic dataset, a challenging real-world traffic monitoring dataset, demonstrate that CA-HQP significantly improves the quality of pseudo-labels compared to existing state-of-the-art cloud-edge collaborative object detection methods. This translates to notable performance gains for the edge model, showcasing the effectiveness of CA-HQP in adapting to dynamic environments. Further ablation studies validate the contribution of each individual component (DQFA, TIAFA, VPG) and confirm the synergistic effect of combining global and instance-level feature alignment strategies. The results highlight the importance of adaptive cloud model updates and sophisticated domain adaptation techniques for achieving robust and accurate object detection in continuously evolving scenarios. The proposed CA-HQP algorithm provides a promising solution for enhancing the performance and reliability of cloud-edge collaborative object detection systems in real-world applications.
\end{abstract}

\begin{IEEEkeywords}
IoT, Cloud-Edge Collaboration,
Object Detection,
Pseudo-Label Generation,
Visual Prompt Tuning
\end{IEEEkeywords}

\section{Introduction}

The escalating demands of urbanization have led to a surge in traffic, exacerbating issues like road congestion and traffic accidents globally \cite{Sadhukhan2021Road}.  Concurrently, the rise of artificial intelligence (AI) and the Internet of Things (IoT) has fostered a paradigm shift in intelligent transportation systems, moving tasks like object detection for traffic monitoring from traditional cloud computing to edge computing  \cite{satyanarayanan2017emergence,ding2022internet}. However, this transition presents a challenge: edge computing environments often rely on lightweight models with limited generalization capabilities \cite{shubha2023adainf}, whereas the dynamic nature of traffic environments necessitates highly generalizable models \cite{gu2022traffic}.
A prominent solution to this challenge lies in the adoption of a cloud-edge collaborative architecture \cite{khani2021real,ding2020cloud,li2019rilod,nan2023large,ding2022emergency}. This framework utilizes lightweight detection models on edge servers for real-time object detection.  Periodically, these servers sample data and transmit it to a central cloud server. The cloud server, equipped with substantial computational resources, utilizes a large, highly generalizable model to generate high-quality pseudo-labels for this unlabeled data. Subsequently, a new model is retrained using the pseudo-labeled data and deployed back to the edge server, completing a retraining cycle. This cycle repeats periodically to enhance model performance \cite{xu2024multi}. 

While previous object detection methods employing cloud-edge collaborative architectures \cite{nan2023large} often assume a perfectly reliable cloud model, this assumption may not always hold true. Although large vision models demonstrate significant generalization capabilities,  fully encompassing the diverse variations inherent in traffic environments during pre-training remains challenging. Consequently, when data sampled from the edge significantly deviates from the cloud model's training distribution, the cloud model can generate erroneous predictions \cite{gan2023cloud}. 
To mitigate this issue, this paper proposes leveraging key data collected from edge devices to adapt the large cloud model before pseudo-label generation.  This adaptation aims to generalize the cloud model to the target domain represented by the sampled edge data. However, the unlabeled nature of this uploaded data precludes supervised fine-tuning. Therefore, we introduce an unsupervised approach to effectively update the cloud model.  Furthermore, we employ DETR (Detection Transformer) \cite{jia2022visual} models for pseudo-label generation. DETR models are the first to offer an end-to-end solution for object detection, achieving state-of-the-art performance in various scenarios \cite{gan2023decorate,lian2024cloud}, making them well-suited for this task. 

This paper first formally defines the problem of cloud-side pseudo-label generation within the context of cloud-edge collaborative object detection. Using the DETR model family as a case study, we briefly introduce their fundamental architecture and identify the challenges associated with domain shifts between the cloud and edge data. To address these challenges, we propose a novel model domain adaptation method based on visual prompt tuning, aiming to enhance the quality of pseudo-label generation. Finally, we present a high-quality pseudo-label generation algorithm that integrates this visual prompt tuning approach. 

\section{Related Work}

CA-HQP draws upon and contributes to several key research areas: cloud-edge collaborative object detection, domain adaptation for object detection, visual prompt tuning, and pseudo-labeling methods.  We position our work within this landscape, highlighting its novel contributions.

\subsection{Cloud-Edge Collaborative Object Detection}

Cloud-edge collaboration has emerged as a promising solution for deploying resource-intensive AI models, such as object detectors, on edge devices with limited computational capabilities \cite{satyanarayanan2017emergence, ali2020res}.  Several works have explored different strategies for collaborative object detection. Shoggoth \cite{wang2023shoggoth} and LVACCL \cite{nan2023large} propose frameworks for adaptive object detection in cloud-edge settings, focusing on efficient task offloading and resource allocation.  However, they often assume a perfectly reliable cloud model for pseudo-label generation, neglecting the potential for domain shift. DCC \cite{gan2023cloud} acknowledges the need for cloud model adaptation and introduces a method based on pixel-level visual prompts and knowledge distillation.  Similarly, EdgeMA \cite{wang2023edgema} proposes a model adaptation system using knowledge distillation and update compression \cite{chen2022update}. However, directly manipulating pixel values can be less effective than feature-level adaptation, and knowledge distillation may not fully capture the target domain's specific characteristics. Other approaches, like Ekya \cite{bhardwaj2022ekya} and CASVA \cite{zhang2022casva}, focus on continuous learning and adaptive model streaming but lack explicit mechanisms for handling domain discrepancies. In contrast, CA-HQP explicitly addresses domain shift by adapting the cloud model to the target domain using a more robust feature-level adaptation strategy guided by learnable visual prompts.  Moreover, CA-HQP considers the practical constraints of bandwidth and latency in cloud-edge systems \cite{li2021appealnet, liu2021petri, liu2022sniper}, striving for efficient and timely adaptation.

\subsection{Domain Adaptation for Object Detection}

Domain adaptation techniques aim to bridge the gap between different data distributions, improving the generalization performance of models across domains \cite{oza2023unsupervised}.  Unsupervised domain adaptation (UDA), where the target domain data is unlabeled, is particularly relevant for cloud-edge collaboration.  Existing UDA methods for object detection often employ adversarial learning \cite{ganin2016domain} to align feature distributions.  For instance, SFA \cite{wang2021exploring} proposes domain query-based feature alignment, achieving promising results. However, these methods primarily focus on global feature alignment and may not effectively address local or instance-level discrepancies. CA-HQP extends these approaches by incorporating both global domain query alignment (DQFA) and a novel instance-aware feature embedding alignment (TIAFA) strategy, ensuring a more comprehensive adaptation to the target domain.  This multi-level alignment, combined with the efficiency of visual prompt tuning, allows CA-HQP to effectively address the domain shift problem in cloud-edge collaborative object detection.

\subsection{Visual Prompt Tuning for Object Detection}

Visual prompt tuning has recently gained traction as an efficient alternative to full fine-tuning for adapting pre-trained vision models \cite{jia2022visual, wang2023review}.  By introducing small learnable parameters (prompts) into the model, prompt tuning can effectively steer the model's behavior without modifying its pre-trained weights.  Existing methods primarily explore prompt tuning in image classification tasks.  In object detection, applying prompts effectively remains a challenge.  Some approaches directly modify input pixel values \cite{gan2023cloud, gan2023decorate}, which can be sensitive to image perturbations.  CA-HQP introduces a novel Visual Prompt Generator (VPG) that operates at the feature level, generating image-specific prompts that are more robust and adaptable.  Furthermore, the use of a learnable prompt generator, inspired by coda-prompt \cite{smith2023coda}, allows for more flexible and efficient adaptation compared to methods using fixed prompts.

\section{High-Quality Pseudo-Label Generation Algorithm based on Visual Prompt Assisted Cloud Model Update}

\begin{figure*}[!h]
    \centering
    \includegraphics[width=1.0\linewidth]{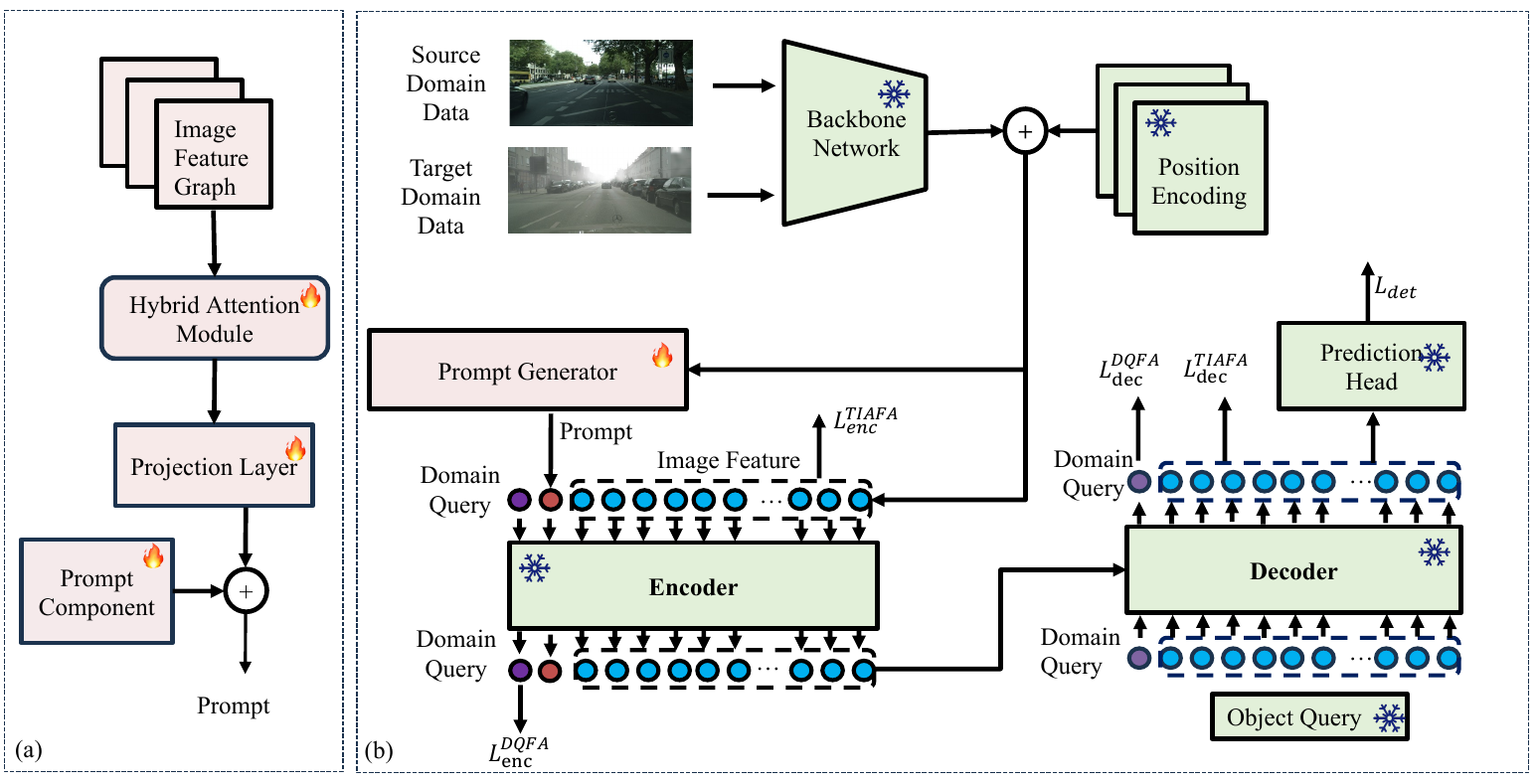}
    \caption{Cloud Model Update with Visual Prompt Generator (VPG) and Feature Alignment. (a) The VPG extracts crucial local features from the input image and generates an image-specific visual prompt. (b) The generated prompt is incorporated into the encoder of the DETR model, along with the image features, facilitating domain adaptation through feature alignment strategies.}
    \label{fig:云端模型更新示意图}
\end{figure*}

Drawing inspiration from previous research in visual prompt tuning \cite{jia2022visual,gan2023decorate}, this paper proposes a novel cloud model adaptation method, as illustrated in Figure \ref{fig:云端模型更新示意图}. This method leverages learnable visual prompts to enhance the flexibility of the cloud model and introduces a multi-granularity feature adversarial loss to facilitate effective domain adaptation.

\subsection{Formalization of the Cloud-Side Pseudo-Label Generation Problem}

Considering the typical architecture of cloud-edge collaboration \cite{khani2021real,ding2020cloud,li2019rilod,nan2023large}, at time step $t$, the edge uploads new data $D_t^{Tar} = \{x_{j,t}^t\}_{j=1}^{N_t}$, where $x_{j,t}^t$ represents the $j$-th data point. The cloud model $f_t$ generates pseudo-labels for $D_t^{Tar}$:

\begin{equation}
\hat{y}_{j,t}^t = \arg\max f_t(x_{j,t}^t).
\end{equation}

Due to the potential distribution discrepancy between $D_t^{Tar}$ and the cloud model's pre-training data $D_S$ \cite{gan2023cloud}, the quality of pseudo-labels might degrade. To enhance the quality, unsupervised domain adaptation techniques \cite{oza2023unsupervised} are employed, aiming to minimize the following objective function to adapt $f_t$:

\begin{equation} \label{eq:问题定义的目标函数}
\min_{f_t} \mathcal{L}_S(D_S, f_t) + \mathcal{D}(D_S, D_t^{Tar}, f_t),
\end{equation}
where $\mathcal{L}_S$ denotes the source domain loss function, and $\mathcal{D}$ represents the inter-domain distance.

\subsection{Learnable Visual Prompt Generator}

To efficiently adjust model parameters, this paper proposes a lightweight Visual Prompt Generator (VPG) as shown in Figure \ref{fig:云端模型更新示意图}(a).

The VPG workflow is as follows:
1) It utilizes a Convolutional Block Attention Module (CBAM) \cite{woo2018cbam} to extract crucial local features from the input image feature map $X \in R^{H \times W \times C}$.
2) The new feature map is then mapped into a vector $q \in R^{1 \times D_p}$, where $D_p$ represents the prompt dimension.
3) Inspired by coda-prompt \cite{smith2023coda}, a learnable visual prompt component $VPC \in R^{D_p \times M}$ ($M$ being the number of components) is employed to obtain the final prompt $\boldsymbol{p}$ through an attention-based weighted sum:

    \begin{equation}
        \boldsymbol{p}= softmax(q \cdot VPC^T) \cdot VPC
    \end{equation}

As depicted in Figure \ref{fig:云端模型更新示意图}(b), the generated visual prompt $\boldsymbol{p}$ participates in the forward process of the encoder module along with the sample:

\begin{align} \label{eq:enc-p}
    [p_i, E_i] & = L_i^{enc}([p_{i-1}, E_{i-1}]) 
\end{align}
where $p_i$ and $E_i$ denote the aggregated visual prompt feature embedding and the image feature embedding computed by the $i$-th encoder module layer, respectively.

To ensure stable prompt updates, an Exponential Moving Average (EMA) strategy is adopted for updating the visual prompt component $VPC$:

\begin{equation}
    VPC_t \leftarrow \beta VPC_{t-1} + (1-\beta) VPC_{t}
\end{equation}
where $\beta$ controls the update magnitude.

\subsection{Domain Adaptation Framework for Cloud Model Update Based on Visual Prompts}

\begin{figure}[!h]
\centering
\includegraphics[width=\linewidth]{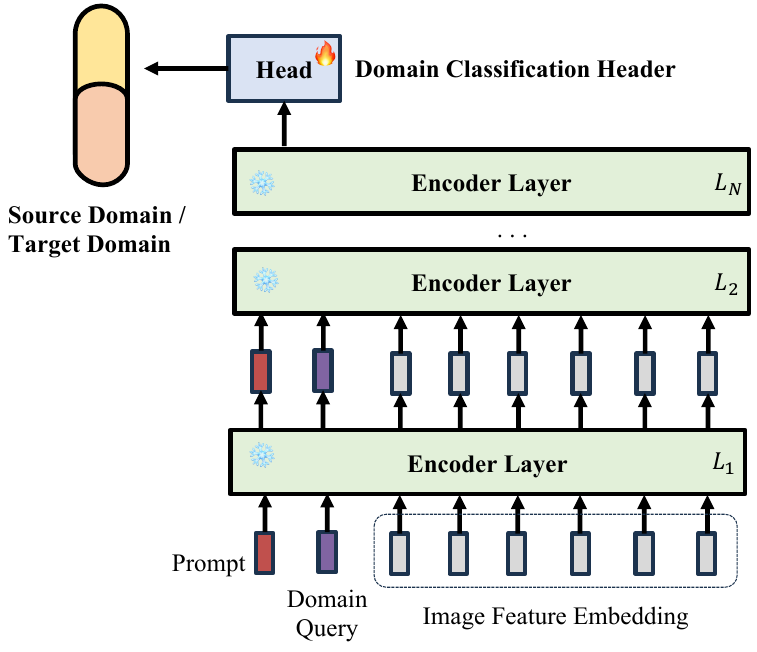}
\caption{Domain adaptation of the cloud model using visual prompts and domain queries.}
\label{fig:pg-detr}
\end{figure}

To adapt the model to new data domains, we propose multi-level cross-domain feature alignment strategies. Inspired by SFA \cite{wang2021exploring}, we first employ domain query to align context features at the global image level (Figure \ref{fig:pg-detr}).

Concisely, the model uses domain query embeddings and visual prompts for domain-invariant feature extraction. The encoder receives both domain query and visual prompts, aiming to extract domain-invariant features that can "fool" two domain discriminators. This adversarial domain adaptation \cite{ganin2016domain} process utilizes gradient reversal layers for parameter updates. Unlike methods directly fine-tuning parameters, this model leverages a Visual Prompt Generator (VPG) to generate image-specific prompts, enhancing feature encoding and generalization.

The ensuing equation details the encoder calculation incorporating visual prompts and domain queries:

\begin{align} \label{eq:enc-p-d}
[p_i, d_i^{enc}, E_i] & = L_i^{enc}([p_{i-1},d_{i-1}^{enc}, E_{i-1}]) \\
y_{d^{enc}} & = head(d_N^{enc})
\end{align}
where $p_i$, $d_i^{enc}$, and $E_i$ represent the aggregated prompt, domain query, and image features from the $i$-th encoder layer, respectively. Here, $L_i^{enc}$ denotes the $i$-th encoder layer, $head(\cdot)$ is the domain discriminator, and $N$ is the number of encoder layers. The decoder module follows a similar structure.

The domain query adversarial loss consists of encoder and decoder components:

\begin{align}
\mathcal{L}{\text {enc}}^{D Q F A} &= t\log D_{enc}^a(y_{d^{enc}})+(1-t) \log (1-D_{enc}^a(y_{d^{enc}})) \label{eq:da-encloss}\\
\mathcal{L}{\text {dec}}^{D Q F A} &= t\log D_{enc}^a(y_{d^{dec}})+(1-t) \log (1-D_{enc}^a(y_{d^{dec}})) \label{eq:da-decloss}
\end{align}
where $\mathcal{D}{enc}^a$ and $\mathcal{D}{dec}^a$ are the domain discriminators for the encoder and decoder, respectively. $t$ represents the domain label (0 for source, 1 for target).

The total domain query adversarial loss is then:

\begin{equation}\label{eq:da-advloss}
\mathcal{L}_{adv}^{DQFA}=\lambda_{1}^{DQFA} \mathcal{L}_{\text {enc }}^{D Q F A}+\lambda_{2}^{DQFA} \mathcal{L}_{\text {dec }}^{D Q F A}
\end{equation}
where $\lambda_1^{DQFA}$ and $\lambda_2^{DQFA}$ are balancing hyperparameters.

To address limitations of global feature alignment in handling local discrepancies, this study proposes an instance-level weighted approach. This approach leverages the DETR model's architecture to identify feature embeddings representing target instances. Specifically, object queries corresponding to sample instances are identified via label assignment.  Subsequently, a similarity matrix between object query embeddings ($Q \in R^{M \times C}$, where $M$ is the number of matched object queries and $C$ is the object query dimension) and encoder output ($z \in R^{C \times d}$, where $d$ is the feature embedding dimension) is used to construct a soft mask ($\psi \in R^d$) for selecting instance object features.

\begin{align} \label{eq:特征选择掩码}
\psi &= \frac{1}{M}\sum_{i=1}^M \boldsymbol{z}^T \cdot \boldsymbol{Q}_i
\end{align}
where $Q_i \in R^{C \times 1}$ represents the $i$-th matched object query.

Since the target domain lacks labels, we employ pseudo-label filtering (threshold $\tau$) to select only high-confidence predictions and obtain a set of reliable pseudo-labels. Equation \ref{eq:特征选择掩码} then yields the feature selection mask. 

This mask selects features for input to the domain discriminator. The adversarial loss for the $i$-th feature embedding in the encoder output is: 

\begin{equation} \label{eq:enc-instance-adv}
\begin{aligned}
\mathcal{L}_{enc}^{TIAFA} = 
& -\frac{1}{N_{enc}} 
\sum_{i=1}^{N_{enc}} 
\left[ t \log D_{enc} 
\left(\psi \odot z_i \right) \right. \\
& \left. + (1 - d) \log \left(1 - D_{enc} \left(\psi \odot z_i \right)\right) \right]
\end{aligned}
\end{equation}
where $z_i$ represents the feature embedding corresponding to the last layer output of the encoder for image features, and $N_{enc}$ represents the number of feature embeddings.

Feature alignment for the decoder module is similar to the encoder, but we only select feature embeddings matching foreground objects. The weighting scheme is as follows:

\begin{equation} \label{eq:解码器模块token权重}
w_{i} =
\begin{cases}
1, & \text{if } q_i \in \mathcal{Y}; \\
0, & \text{else}.
\end{cases}
\end{equation}
where $q_i$ represents the $i$-th output of the last layer of the decoder corresponding to the object query, and $\mathcal{Y}$ represents the set of object query outputs matched with annotations.

The feature embedding alignment function for the decoder module is:

\begin{equation} \label{eq:dec-instance-adv}
\begin{aligned}
\mathcal{L}_{dec}^{TIAFA} = & -\frac{1}{N_{dec}} \sum_{i=1}^{N_{dec}} \left[ t \log D_{dec} \left(w_i \cdot q_i \right) \right. \\
& \left. + (1 - d) \log \left(1-D_{dec} \left(w_i \cdot q_i \right)\right) \right];
\end{aligned}
\end{equation}
where $N_{dec}$ represents the number of object queries, $q_i$ is the $i$-th feature embedding, and $D_{dec}$ is the corresponding domain discriminator.

The total instance-aware feature embedding alignment loss is:

\begin{equation}
\mathcal{L}_{adv}^{TIAFA} = \lambda_1^{TIAFA} \mathcal{L}_{enc}^{TIAFA} + \lambda_2^{TIAFA} \mathcal{L}_{dec}^{TIAFA}
\end{equation}
where $\lambda_1^{TIAFA}$ and $\lambda_2^{TIAFA}$ are hyperparameters for balancing.

The final optimization objective function is defined as follows:

\begin{equation}\label{eq:da-loss}
\begin{aligned}
\mathcal{L}_{all} &= \mathcal{L}_{\text{det}} - \mathcal{L}_{adv} \\
&= \mathcal{L}_{\text{det}} - (\mathcal{L}_{adv}^{DQFA} + \mathcal{L}_{adv}^{TIAFA})
\end{aligned}
\end{equation}
where $\mathcal{L}_{\text{det}}$ is the detection loss function.

\subsection{High-Quality Pseudo-Label Generation Algorithm Based on Cloud Model Adaptation}

We propose a visual prompt-based high-quality pseudo-label generation algorithm (Algorithm \ref{alg:ca-hqp}). This algorithm generates visual prompts for target domain images using the VPG, applies these prompts to the model's encoder and decoder, updates model and VPG parameters by minimizing the combined detection and adversarial losses, and finally generates high-quality pseudo-labels using the updated model.

\begin{algorithm}
\caption{\textbf{High-Quality Pseudo-Label Generation Algorithm Based on Visual Prompt Assisted Cloud Model Update (CA-HQP)}}
\label{alg:ca-hqp}
\begin{algorithmic}[1]
\Statex \hspace*{-\algorithmicindent} Input: Source domain dataset $\mathcal{D}_{S}$, Target domain dataset $\mathcal{D}_{t}^T$, Cloud model $f_t$, Visual Prompt Generator $VPG$, Domain discriminator $D$
\Statex \hspace*{-\algorithmicindent} Output: Updated model adapted to the target domain $f_{t+1}$, Pseudo-labels for the target domain $\hat{Y}$
\For{each training epoch}
    \For{each batch of data $I_{S} \in \mathcal{D}_{S}$ and $I_{t}^T \in \mathcal{D}_{t}^T$}
        \State Generate visual prompt $p$ for $I_{t}^T$ using $VPG$
        \State Apply $p$ to the encoder and decoder modules of $f_t$
        \State Perform forward propagation of $f_t$ using $I_{S}$ and $I_{t}^T$
        \State Calculate the detection loss $\mathcal{L}_{\text{det}}$ using $I_{S}$ 
        \State Calculate the adversarial learning loss $\mathcal{L}_{\text{adv}}$ using the domain discriminator $D$
        \State Calculate the total loss $\mathcal{L}_{\text{all}} = \mathcal{L}_{\text{det}} - \mathcal{L}_{\text{adv}}$
        \State Backpropagate and update the parameters of $f_t$ and $VPG$
    \EndFor
\EndFor

\State Initialize the pseudo-label set $\hat{{Y}}$
\For{each unlabeled target domain data $I_{t}^T$}
    \State Use the updated model $f_{t+1}$ to make predictions on $I_{t}^T$
    \State Generate pseudo-label $\hat{y}$ and add it to $\hat{{Y}}$
\EndFor

\State \Return Updated model $f_{t+1}$, Generated pseudo-label set $\hat{{Y}}$
\end{algorithmic}
\end{algorithm}

\section{Experiments}

This section validates the effectiveness of the proposed Cloud-Adaptive High-Quality Pseudo-label generation algorithm (CA-HQP) for continuous learning in dynamic traffic monitoring scenarios. We evaluate CA-HQP by integrating it into existing cloud-edge collaborative object detection frameworks and comparing their performance on the Bellevue traffic video dataset.

\subsection{Experimental Setup}

\paragraph{Dataset} The Bellevue traffic video dataset comprises footage from 8 distinct cameras capturing real-world traffic scenes. Each video sequence is approximately 30 minutes long with a frame rate of 30 FPS, resulting in roughly 54,000 frames per sequence. The dataset presents challenges like varying lighting conditions (day/night), weather changes, and diverse traffic densities. Annotations include bounding boxes for vehicles with class labels (car, bus, truck, etc.).  For each experiment, the first half of the video sequence is used for training (domain adaptation), and the second half is used for testing.

\paragraph{Implementation Details}  All models are implemented using the PyTorch deep learning framework and trained on a server equipped with 4 NVIDIA RTX 3090 GPUs.  The cloud server utilizes the DINO object detection model with a ResNet-101 backbone pre-trained on ImageNet. The edge server employs the RT-DETR model with a ResNet-18 backbone, chosen for its real-time performance capabilities.

During training, we use the AdamW optimizer with a learning rate of $1e^{-4}$ for the cloud model and $1e^{-3}$ for the VPG. The batch size is set to 16, and training is performed for 100 epochs.  We apply standard data augmentation techniques like random horizontal flipping, cropping, and color jittering to improve generalization. Hyperparameters for CA-HQP ($\beta$, $\lambda_1^{DQFA}$, $\lambda_2^{DQFA}$, $\lambda_1^{TIAFA}$, $\lambda_2^{TIAFA}$) are determined through grid search on a validation set, optimizing for edge model performance.

\paragraph{Baselines} We compare CA-HQP against the following state-of-the-art cloud-edge collaborative object detection methods:

\begin{itemize}
    \item \textbf{Shoggoth}\cite{wang2023shoggoth}: A cloud-edge collaborative adaptive object detection method that assumes a perfectly reliable cloud model and does not consider updating the cloud model during pseudo-label generation.
    \item \textbf{Shoggoth-CA-HQP}: This variant replaces the pseudo-label generation algorithm in Shoggoth with the proposed CA-HQP algorithm.
    \item \textbf{DCC}\cite{gan2023cloud}:  A cloud-edge collaborative adaptive object detection method that, for the first time, considers updating the cloud model during pseudo-label generation. It introduces pixel-level visual prompts to facilitate model adaptation.
    \item \textbf{DCC-CA-HQP}: This variant replaces the pseudo-label generation algorithm in DCC with the proposed CA-HQP algorithm.
    \item \textbf{LVACCL}\cite{nan2023large}:  A framework leveraging a large-scale pre-trained cloud model for generating high-quality pseudo-labels.
    \item \textbf{LVACCL-CA-HQP} This variant replaces the pseudo-label generation algorithm in LVACCL with the proposed CA-HQP algorithm.
\end{itemize}

\begin{table*}[!h]
    \centering
    \caption{Validity Verification of CA-HQP by Pseudo-Label.} \label{tab:CA-HQP算法有效性验证-伪标签准确性}
    \resizebox{0.9\linewidth}{!}{
    \begin{tabular}{lccccccccc} 
    \hline
    Time & \multicolumn{8}{c}{t$\xrightarrow{\rule{0.4\linewidth}{0pt}}$} &\\
    \hline
    Video Sequence & 1 & 2 & 3 & 4 & 5 & 6 & 7 & 8 & Mean\\
    \hline
    Shoggoth & 60.93 & 55.57 & 57.22 & 55.36 & 66.41 & 63.71 & 64.04 & 56.49 & 59.97\\
    Shoggoth-CA-HQP & 61.29 & 58.49 & 58.59 & 60.36 & 68.17 & 65.81 & 66.28 & 60.65 & 62.45\\
    DCC & 62.21 & 56.39 & 58.06 & 59.86 & 69.39 & 67.26 & 63.74 & 61.09 & 62.25\\
    DCC-CA-HQP & 62.71 & 57.58 & 58.84 & 60.57 & 71.34 & 68.81 & 64.95 & 62.63 & \underline{\textbf{63.26}}\\
    LVACCL & 60.61 & 55.76 & 56.95 & 55.26 & 65.8 & 63.54 & 60.12 & 57.15 & 59.40\\
    LVACCL-CA-HQP & 61.96 & 57.64 & 57.48 & 58.47 & 67.84 & 66.21 & 63.38 & 58.97 & 61.63\\
    \hline
    \end{tabular}
    }
\end{table*}

\begin{table*}[!h]
    \caption{Validity Verification of CA-HQP by Edge Model.}
    \centering
    \resizebox{0.9\linewidth}{!}{
    \label{tab:CA-HQP算法有效性验证-边缘端模型的准确性}
    \begin{tabular}{lccccccccc} 
    \hline
    Time & \multicolumn{8}{c}{t$\xrightarrow{\rule{0.4\linewidth}{0pt}}$} & \\
    \hline
    Video Sequence & 1 & 2 & 3 & 4 & 5 & 6 & 7 & 8 & Mean\\
    \hline
    
    Shoggoth & 57.67 & 53.22 & 53.69 & 51.77 & 63.18 & 60.56 & 59.66 & 52.74 & 56.56\\
    Shoggoth-CA-HQP & 57.87 & 54.50 & 55.24 & 56.87 & 64.17 & 61.87 & 62.19 & 56.73 & 58.68\\
    DCC & 59.16 & 53.28 & 54.47 & 57.16 & 66.43 & 64.00 & 60.09 & 58.55 & 59.14\\
    DCC-CA-HQP & 59.96 & 53.49 & 54.78 & 57.74 & 68.33 & 64.49 & 60.87 & 58.87 & \underline{\textbf{59.82}}\\
    LVACCL & 57.20 & 52.09 & 53.59 & 52.03 & 61.94 & 60.55 & 56.99 & 54.66 & 56.13\\
    LVACCL-CA-HQP & 57.48 & 54.87 & 54.24 & 55.97 & 65.50 & 62.76 & 60.26 & 55.51 & 58.32\\
    \hline
    \end{tabular}
    }
\end{table*}

\subsection{Performance Validation of the CA-HQP} 


\begin{table*}[!h]
\caption{CA-HQP Ablation Study with Sloggoth.}
\centering
\label{tab:CA-HQP算法消融研究-Shoggoth}
\resizebox{0.65\linewidth}{!}{
\begin{tabular}{ccccc}
\toprule
\multicolumn{1}{l}{DQFA} & TIAFA & VPG & \multicolumn{1}{l}{Pseudo-Label Acc} & Edge Model Detection Acc \\
\midrule
X & X & X & 59.97 & 56.56  \\
\checkmark & X & X & 60.11 & 56.85 \\
X & \checkmark & X & 60.51 & 56.99 \\
\checkmark & \multicolumn{1}{c}{\checkmark} & X & 61.8 & 57.75 \\
\checkmark & \multicolumn{1}{c}{\checkmark} & \multicolumn{1}{c}{\checkmark} & \underline{\textbf{62.45}} & \underline{\textbf{58.68}} \\
\bottomrule
\end{tabular}
}
\end{table*}

\begin{table*}[!h]
\caption{CA-HQP Ablation Study with DCC.}
\centering
\resizebox{0.65\linewidth}{!}{
\label{tab:CA-HQP算法消融研究-DCC}
\begin{tabular}{ccccc}
\toprule
\multicolumn{1}{l}{DQFA} & TIAFA & VPG & \multicolumn{1}{l}{Pseudo-Label Acc} & Edge Model Detection Acc \\
\midrule
X & X & X & 62.25 & 59.14  \\
\checkmark & X & X & 62.39 & 59.21 \\
X & \checkmark & X & 62.51 & 59.40 \\
\checkmark & \multicolumn{1}{c}{\checkmark} & X & 62.81 & 59.60 \\
\checkmark & \multicolumn{1}{c}{\checkmark} & \multicolumn{1}{c}{\checkmark} & \underline{\textbf{63.26}} & \underline{\textbf{59.82}} \\
\bottomrule
\end{tabular}
}
\end{table*}

\begin{table*}[!h]
\caption{CA-HQP Ablation Study with LVACCL.}
\label{tab:CA-HQP算法消融研究-LVACCL}
\centering
\resizebox{0.65\linewidth}{!}{
\begin{tabular}{ccccc}
\toprule
\multicolumn{1}{l}{DQFA} & TIAFA & VPG & \multicolumn{1}{l}{Pseudo-Label Acc} & Edge Model Detection Acc \\
\midrule
X & X & X & 59.40 & 56.13  \\
\checkmark & X & X & 59.91 & 56.85 \\ 
X & \checkmark & X & 60.12 & 57.04 \\
\checkmark & \multicolumn{1}{c}{\checkmark} & X & 60.88 & 57.99 \\
\checkmark & \multicolumn{1}{c}{\checkmark} & \multicolumn{1}{c}{\checkmark} & \underline{\textbf{61.63}} & \underline{\textbf{58.82}} \\
\bottomrule
\end{tabular}
}
\end{table*}




\paragraph{Performance Analysis} Tables \ref{tab:CA-HQP算法有效性验证-伪标签准确性} and \ref{tab:CA-HQP算法有效性验证-边缘端模型的准确性} present a comprehensive performance comparison of CA-HQP against state-of-the-art cloud-edge collaborative object detection methods on the Bellevue traffic dataset. The evaluation metric used is mean Average Precision (mAP), a standard measure for object detection accuracy.

As shown in Table \ref{tab:CA-HQP算法有效性验证-伪标签准确性}, CA-HQP consistently improves the quality of generated pseudo-labels across all baseline methods. For instance, Shoggoth achieves a pseudo-label mAP of 59.97\%, while incorporating CA-HQP (Shoggoth-CA-HQP) boosts the mAP to 62.45\%, a significant improvement of 2.48\%. Similarly, DCC, which already incorporates cloud model updates, sees a further improvement from 62.25\% to 63.26\% with CA-HQP. This highlights the effectiveness of CA-HQP's visual prompt and feature alignment mechanisms in enhancing the cloud model's adaptation to the target domain.

The impact of improved pseudo-label quality translates directly to enhanced edge model performance, as evidenced in Table \ref{tab:CA-HQP算法有效性验证-边缘端模型的准确性}. The DCC-CA-HQP variant achieves the highest edge model mAP of 59.82\%, outperforming all other methods. Notably, CA-HQP provides substantial gains for baselines that rely on static cloud models (Shoggoth and LVACCL), demonstrating its ability to compensate for the lack of cloud model adaptation in these methods. For example, LVACCL's edge mAP increases from 56.13\% to 58.32\% with the integration of CA-HQP. Across all video sequences, the methods incorporating CA-HQP exhibit more stable and consistently higher performance compared to their counterparts without CA-HQP, indicating better robustness to the dynamic nature of the traffic scenes.

\paragraph{Ablation Study} Tables \ref{tab:CA-HQP算法消融研究-Shoggoth}, \ref{tab:CA-HQP算法消融研究-DCC}, and \ref{tab:CA-HQP算法消融研究-LVACCL} present the results of the ablation study, which investigates the contribution of each component within CA-HQP: DQFA, TIAFA, and VPG.

Across all baselines, removing any single component leads to a performance drop, confirming their importance. TIAFA consistently demonstrates a larger individual impact compared to DQFA. For example, in the Shoggoth ablation (Table \ref{tab:CA-HQP算法消融研究-Shoggoth}), using only TIAFA results in a pseudo-label mAP of 60.51\%, while using only DQFA yields 60.11\%. However, the combination of DQFA and TIAFA significantly outperforms using either alone, indicating a synergistic effect. This suggests that global and instance-level feature alignments are complementary and contribute to a more comprehensive domain adaptation.

The VPG plays a crucial role in achieving parameter-efficient domain adaptation. Removing the VPG (equivalent to full fine-tuning) consistently reduces performance. For instance, in DCC (Table \ref{tab:CA-HQP算法消融研究-DCC}), removing the VPG drops the pseudo-label mAP from 63.26\% to 62.81\%, demonstrating that the VPG enables better adaptation with fewer trainable parameters. This suggests that the VPG effectively guides the adaptation process, focusing on task-relevant adjustments while preserving the pre-trained model's general knowledge.

\section{Conclusion}
This paper proposed CA-HQP, a novel algorithm for generating high-quality pseudo-labels in cloud-edge collaborative object detection for dynamic traffic monitoring scenarios. CA-HQP addresses the limitations of existing methods by incorporating a learnable Visual Prompt Generator (VPG) for parameter-efficient cloud model adaptation and a dual feature alignment strategy comprising global Domain Query Feature Alignment (DQFA) and instance-aware Temporal Instance-Aware Feature Embedding Alignment (TIAFA). This approach enables the cloud model to effectively adapt to the evolving target domain data distribution, resulting in more accurate and reliable pseudo-labels.
Extensive experiments conducted on the Bellevue traffic dataset demonstrate that CA-HQP significantly improves both the quality of generated pseudo-labels and the performance of the edge model. CA-HQP consistently outperforms existing state-of-the-art cloud-edge collaborative object detection methods, showcasing its effectiveness in handling the challenges of dynamic traffic scenes. The quantitative results highlight the superior performance of CA-HQP in terms of both pseudo-label accuracy and edge model detection mAP. Furthermore, ablation studies confirm the contribution of each individual component within CA-HQP, namely the VPG, DQFA, and TIAFA. The results emphasize the synergistic benefits of combining global and instance-level feature alignment strategies for achieving comprehensive domain adaptation. The VPG proves crucial for enabling efficient adaptation by minimizing the number of trainable parameters while maximizing performance gains.

\paragraph*{Future Work}

While CA-HQP demonstrates promising results, several avenues for future research exist. Investigating alternative visual prompt engineering techniques could further enhance the adaptability of the cloud model. Exploring different feature alignment strategies or incorporating temporal consistency constraints into the pseudo-label generation process might lead to additional performance improvements. Moreover, extending CA-HQP to other application domains beyond traffic monitoring and evaluating its performance on more diverse datasets would provide further insights into its generalizability and robustness.

\bibliographystyle{IEEEtran}
\bibliography{refs}

\begin{thebibliography}{10}
\providecommand{\url}[1]{#1}
\csname url@samestyle\endcsname
\providecommand{\newblock}{\relax}
\providecommand{\bibinfo}[2]{#2}
\providecommand{\BIBentrySTDinterwordspacing}{\spaceskip=0pt\relax}
\providecommand{\BIBentryALTinterwordstretchfactor}{4}
\providecommand{\BIBentryALTinterwordspacing}{\spaceskip=\fontdimen2\font plus
\BIBentryALTinterwordstretchfactor\fontdimen3\font minus \fontdimen4\font\relax}
\providecommand{\BIBforeignlanguage}[2]{{%
\expandafter\ifx\csname l@#1\endcsname\relax
\typeout{** WARNING: IEEEtran.bst: No hyphenation pattern has been}%
\typeout{** loaded for the language `#1'. Using the pattern for}%
\typeout{** the default language instead.}%
\else
\language=\csname l@#1\endcsname
\fi
#2}}
\providecommand{\BIBdecl}{\relax}
\BIBdecl

\bibitem{Sadhukhan2021Road}
P.~Sadhukhan, S.~Banerjee, and P.~Das, ``Road traffic congestion monitoring in urban areas: A review,'' pp. 199--211, 2021.

\bibitem{satyanarayanan2017emergence}
M.~Satyanarayanan, ``The emergence of edge computing,'' \emph{Computer}, vol.~50, no.~1, pp. 30--39, 2017.

\bibitem{ding2022internet}
Z.~Ding, S.~Jiang, X.~Xu, and Y.~Han, ``An internet of things based scalable framework for disaster data management,'' \emph{Journal of safety science and resilience}, vol.~3, no.~2, pp. 136--152, 2022.

\bibitem{shubha2023adainf}
S.~S. Shubha and H.~Shen, ``Adainf: Data drift adaptive scheduling for accurate and slo-guaranteed multiple-model inference serving at edge servers,'' in \emph{Proceedings of the ACM SIGCOMM 2023 Conference}, 2023, pp. 473--485.

\bibitem{gu2022traffic}
Z.~Gu, Y.~Lei, S.~Chan, D.~Cao, and K.~Zhang, ``Traffic incident detection system based on video analysis,'' in \emph{Proceedings of the 2022 5th International Conference on Computational Intelligence and Intelligent Systems}, 2022, pp. 15--20.

\bibitem{khani2021real}
M.~Khani, P.~Hamadanian, A.~Nasr-Esfahany, and M.~Alizadeh, ``Real-time video inference on edge devices via adaptive model streaming,'' in \emph{Proceedings of the IEEE/CVF International Conference on Computer Vision}, 2021, pp. 4572--4582.

\bibitem{ding2020cloud}
C.~Ding, A.~Zhou, Y.~Liu, R.~N. Chang, C.-H. Hsu, and S.~Wang, ``A cloud-edge collaboration framework for cognitive service,'' \emph{IEEE Transactions on Cloud Computing}, vol.~10, no.~3, pp. 1489--1499, 2020.

\bibitem{li2019rilod}
D.~Li, S.~Tasci, S.~Ghosh, J.~Zhu, J.~Zhang, and L.~Heck, ``Rilod: Near real-time incremental learning for object detection at the edge,'' in \emph{Proceedings of the 4th ACM/IEEE Symposium on Edge Computing}, 2019, pp. 113--126.

\bibitem{nan2023large}
Y.~Nan, S.~Jiang, and M.~Li, ``Large-scale video analytics with cloud--edge collaborative continuous learning,'' \emph{ACM Transactions on Sensor Networks}, vol.~20, no.~1, pp. 1--23, 2023.

\bibitem{ding2022emergency}
Z.~Ding, X.~Xu, S.~Jiang, J.~Yan, and Y.~Han, ``Emergency logistics scheduling with multiple supply-demand points based on grey interval,'' \emph{Journal of Safety Science and Resilience}, vol.~3, no.~2, pp. 179--188, 2022.

\bibitem{xu2024multi}
X.~Xu, Z.~Lian, Y.~Wu, M.~Lv, Z.~Ding, J.~Yan, and S.~Jiang, ``A multi-constraint and multi-objective allocation model for emergency rescue in iot environment,'' in \emph{2024 IEEE International Symposium on Circuits and Systems (ISCAS)}.\hskip 1em plus 0.5em minus 0.4em\relax IEEE, 2024, pp. 1--5.

\bibitem{gan2023cloud}
Y.~Gan, M.~Pan, R.~Zhang, Z.~Ling, L.~Zhao, J.~Liu, and S.~Zhang, ``Cloud-device collaborative adaptation to continual changing environments in the real-world,'' in \emph{Proceedings of the IEEE/CVF Conference on Computer Vision and Pattern Recognition}, 2023, pp. 12\,157--12\,166.

\bibitem{jia2022visual}
M.~Jia, L.~Tang, B.-C. Chen, C.~Cardie, S.~Belongie, B.~Hariharan, and S.-N. Lim, ``Visual prompt tuning,'' in \emph{European Conference on Computer Vision}.\hskip 1em plus 0.5em minus 0.4em\relax Springer, 2022, pp. 709--727.

\bibitem{gan2023decorate}
Y.~Gan, Y.~Bai, Y.~Lou, X.~Ma, R.~Zhang, N.~Shi, and L.~Luo, ``Decorate the newcomers: Visual domain prompt for continual test time adaptation,'' in \emph{Proceedings of the AAAI Conference on Artificial Intelligence}, vol.~37, no.~6, 2023, pp. 7595--7603.

\bibitem{lian2024cloud}
Z.~Lian, M.~Lv, X.~Xu, Z.~Ding, M.~Zhu, Y.~Wu, and J.~Yan, ``Cloud-edge collaborative continual adaptation for its object detection,'' in \emph{International Conference on Spatial Data and Intelligence}.\hskip 1em plus 0.5em minus 0.4em\relax Springer, 2024, pp. 15--27.

\bibitem{ali2020res}
M.~Ali, A.~Anjum, O.~Rana, A.~R. Zamani, D.~Balouek-Thomert, and M.~Parashar, ``Res: Real-time video stream analytics using edge enhanced clouds,'' \emph{IEEE Transactions on Cloud Computing}, vol.~10, no.~2, pp. 792--804, 2020.

\bibitem{wang2023shoggoth}
L.~Wang, K.~Lu, N.~Zhang, X.~Qu, J.~Wang, J.~Wan, G.~Li, and J.~Xiao, ``Shoggoth: towards efficient edge-cloud collaborative real-time video inference via adaptive online learning,'' in \emph{2023 60th ACM/IEEE Design Automation Conference (DAC)}.\hskip 1em plus 0.5em minus 0.4em\relax IEEE, 2023, pp. 1--6.

\bibitem{wang2023edgema}
L.~Wang, N.~Zhang, X.~Qu, J.~Wang, J.~Wan, G.~Li, K.~Hu, G.~Jiang, and J.~Xiao, ``Edgema: Model adaptation system for real-time video analytics on edge devices,'' in \emph{International Conference on Neural Information Processing}.\hskip 1em plus 0.5em minus 0.4em\relax Springer, 2023, pp. 292--304.

\bibitem{chen2022update}
B.~Chen, A.~Bakhshi, G.~Batista, B.~Ng, and T.-J. Chin, ``Update compression for deep neural networks on the edge,'' in \emph{Proceedings of the IEEE/CVF Conference on Computer Vision and Pattern Recognition}, 2022, pp. 3076--3086.

\bibitem{bhardwaj2022ekya}
R.~Bhardwaj, Z.~Xia, G.~Ananthanarayanan, J.~Jiang, Y.~Shu, N.~Karianakis, K.~Hsieh, P.~Bahl, and I.~Stoica, ``Ekya: Continuous learning of video analytics models on edge compute servers,'' in \emph{19th USENIX Symposium on Networked Systems Design and Implementation (NSDI 22)}, 2022, pp. 119--135.

\bibitem{zhang2022casva}
M.~Zhang, F.~Wang, and J.~Liu, ``Casva: Configuration-adaptive streaming for live video analytics,'' in \emph{IEEE INFOCOM 2022-IEEE Conference on Computer Communications}.\hskip 1em plus 0.5em minus 0.4em\relax IEEE, 2022, pp. 2168--2177.

\bibitem{li2021appealnet}
M.~Li, Y.~Li, Y.~Tian, L.~Jiang, and Q.~Xu, ``Appealnet: An efficient and highly-accurate edge/cloud collaborative architecture for dnn inference,'' in \emph{2021 58th ACM/IEEE Design Automation Conference (DAC)}.\hskip 1em plus 0.5em minus 0.4em\relax IEEE, 2021, pp. 409--414.

\bibitem{liu2021petri}
R.~Liu, L.~Zhang, J.~Wang, H.~Yang, and Y.~Liu, ``Petri: Reducing bandwidth requirement in smart surveillance by edge-cloud collaborative adaptive frame clustering and pipelined bidirectional tracking,'' in \emph{2021 58th ACM/IEEE Design Automation Conference (DAC)}.\hskip 1em plus 0.5em minus 0.4em\relax IEEE, 2021, pp. 421--426.

\bibitem{liu2022sniper}
W.~Liu, J.~Geng, Z.~Zhu, J.~Cao, and Z.~Lian, ``Sniper: Cloud-edge collaborative inference scheduling with neural network similarity modeling,'' in \emph{Proceedings of the 59th ACM/IEEE Design Automation Conference}, 2022, pp. 505--510.

\bibitem{oza2023unsupervised}
P.~Oza, V.~A. Sindagi, V.~V. Sharmini, and V.~M. Patel, ``Unsupervised domain adaptation of object detectors: A survey,'' \emph{IEEE Transactions on Pattern Analysis and Machine Intelligence}, 2023.

\bibitem{ganin2016domain}
Y.~Ganin, E.~Ustinova, H.~Ajakan, P.~Germain, H.~Larochelle, F.~Laviolette, M.~March, and V.~Lempitsky, ``Domain-adversarial training of neural networks,'' \emph{Journal of machine learning research}, vol.~17, no.~59, pp. 1--35, 2016.

\bibitem{wang2021exploring}
W.~Wang, Y.~Cao, J.~Zhang, F.~He, Z.-J. Zha, Y.~Wen, and D.~Tao, ``Exploring sequence feature alignment for domain adaptive detection transformers,'' in \emph{Proceedings of the 29th ACM International Conference on Multimedia}, 2021, pp. 1730--1738.

\bibitem{wang2023review}
J.~Wang, Z.~Liu, L.~Zhao, Z.~Wu, C.~Ma, S.~Yu, H.~Dai, Q.~Yang, Y.~Liu, S.~Zhang \emph{et~al.}, ``Review of large vision models and visual prompt engineering,'' \emph{Meta-Radiology}, p. 100047, 2023.

\bibitem{smith2023coda}
J.~S. Smith, L.~Karlinsky, V.~Gutta, P.~Cascante-Bonilla, D.~Kim, A.~Arbelle, R.~Panda, R.~Feris, and Z.~Kira, ``Coda-prompt: Continual decomposed attention-based prompting for rehearsal-free continual learning,'' in \emph{Proceedings of the IEEE/CVF Conference on Computer Vision and Pattern Recognition}, 2023, pp. 11\,909--11\,919.

\bibitem{woo2018cbam}
S.~Woo, J.~Park, J.-Y. Lee, and I.~S. Kweon, ``Cbam: Convolutional block attention module,'' in \emph{Proceedings of the European conference on computer vision (ECCV)}, 2018, pp. 3--19.

\end{thebibliography}

\end{document}